\documentclass[10pt]{allhands}

\definecolor{darkgreen}{RGB}{0,128,0}

\usepackage{amsfonts}
\usepackage{amsmath}
\usepackage{amsthm}
\usepackage{wrapfig}
\usepackage{algpseudocode}
\usepackage[linesnumbered,lined,boxed,commentsnumbered,ruled,longend]{algorithm2e}
\usepackage[capitalize,noabbrev]{cleveref}

\usepackage{tikz}
\usetikzlibrary{positioning,fit,calc}

\usepackage{fancyvrb}
\usepackage{fvextra}

\theoremstyle{plain}

\theoremstyle{definition}

\theoremstyle{remark}

\usepackage{enumitem}
\usepackage{xspace}

\newenvironment{itemize*}%
 {\leftmargini=10pt\begin{itemize}%
  \setlength{\itemsep}{0pt}%
  \setlength{\parskip}{0pt}%
  }%
 {\end{itemize}}
\newenvironment{enumerate*}%
 {\begin{enumerate}%
  \setlength{\itemsep}{0pt}%
  \setlength{\parskip}{0pt}}%
 {\end{enumerate}}

\newcommand{\sref}[1]{\S\ref{#1}}
\newcommand{\fref}[1]{Fig.~\ref{#1}}
\newcommand{\tref}[1]{Tab.~\ref{#1}}

\newcommand{\numProdSegments}{151,837\xspace}        %
\newcommand{\numProdTrain}{150,290\xspace}           %
\newcommand{\numProdTest}{1,547\xspace}              %
\newcommand{\numSweGym}{4,238\xspace}                %
\newcommand{\numTotalTrainK}{154K\xspace}            %
\newcommand{\pctSurvival}{4\%\xspace}                %
\newcommand{\pctMerge}{6\%\xspace}                   %
\newcommand{\pctUnlabeled}{96\%\xspace}              %
\newcommand{\numProdWithSurvival}{5,349\xspace}      %
\newcommand{\numProdWithMerge}{9,750\xspace}         %
\newcommand{\numProdConvs}{38,241\xspace}            %
\newcommand{\meanRubricAUC}{0.78\xspace}             %
\newcommand{\topRubricAUC}{0.81\xspace}              %
\newcommand{\numRubrics}{24\xspace}                  %

\usepackage{color-edits}

\addauthor{gn}{magenta}
\addauthor{vc}{blue}
\newcommand{\methodname}{Critic Rubrics}

\begin{document}

\title{A Rubric-Supervised Critic from Sparse Real-World Outcomes}

\author[1,3]{Xingyao Wang}
\author[2]{Valerie Chen}
\author[3]{Heng Ji}
\author[1,2]{Graham Neubig}

\contact{\{xingyao,graham\}@openhands.dev}

\affiliation[1]{OpenHands}
\affiliation[2]{CMU}
\affiliation[3]{UIUC}

\abstract{%
Academic benchmarks for coding agents tend to reward \emph{autonomous} task completion, measured by \emph{verifiable rewards} such as unit-test success. In contrast, real-world coding agents operate \emph{with humans in the loop}, where success signals are typically \emph{noisy, delayed, and sparse}.
How can we bridge this gap?
In this paper, we propose a process to learn a ``critic'' model from sparse and noisy interaction data, which can then be used both as a reward model for either RL-based training or inference-time scaling.
Specifically, we introduce \methodname, a rubric-based supervision framework with \numRubrics behavioral features that can be derived from human-agent interaction traces alone.
Using a semi-supervised objective, we can then jointly predict these rubrics and sparse human feedback (when present).
In experiments, we demonstrate that, despite being trained primarily from trace-observable rubrics and sparse real-world outcome proxies, these critics improve best-of-N reranking on SWE-bench (Best@8 +15.9 over Random@8 over the rerankable subset of trajectories), enable early stopping (+17.7 with 83\% fewer attempts), and support training-time data curation via critic-selected trajectories.

}

\metadata[Critic Rubrics]{\url{https://github.com/OpenHands/critic-rubrics}}
\metadata[Model]{\url{https://huggingface.co/OpenHands/openhands-critic-4b-v1.0}}
\metadata[Docs]{\url{https://docs.openhands.dev/sdk/guides/critic}}

\maketitle

\section{Introduction}
\label{sec:introduction}

\begin{figure*}[t]
  \centering
  \includegraphics[width=\textwidth]{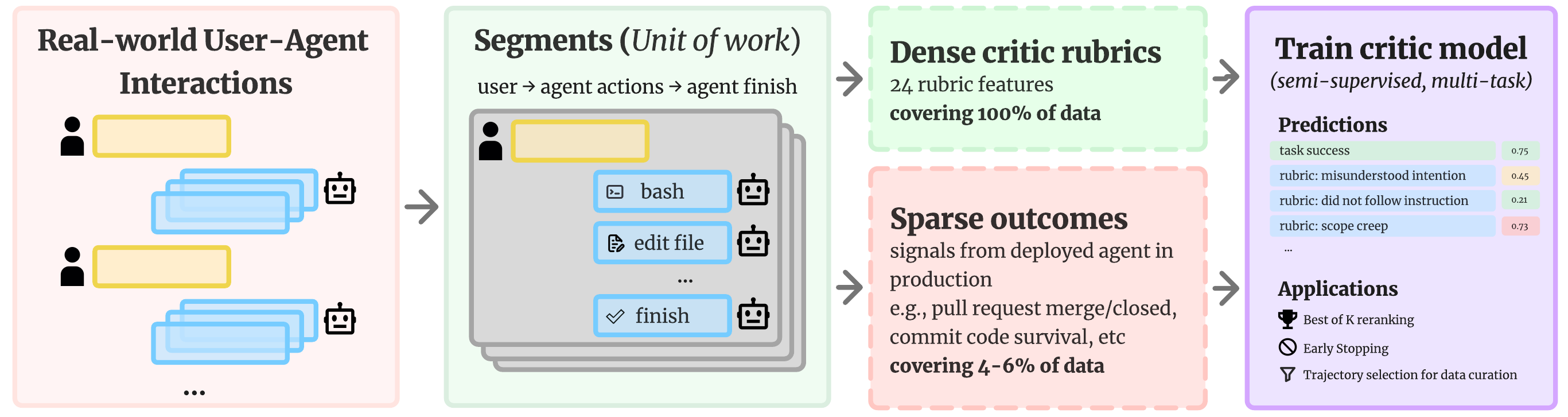}
  \vspace{-15px}
  \caption{\textbf{Overview of our method: Learning a deployable critic from production traces.} We convert real-world human--agent interactions into segments (user request $\rightarrow$ agent actions $\rightarrow$ finish), annotate each segment with trace-observable Critic Rubrics (24 dense behavioral signals), and combine them with sparse production outcome proxies (e.g., PR merge / code survival) to train a semi-supervised, multi-task critic that predicts both rubric features and segment success. The resulting critic supports best-of-$K$ reranking, compute-efficient early stopping, and trajectory selection for training-time data curation.}
  \label{fig:overview}
  \vspace{-0.8em}
\end{figure*}

Today, LLM-powered software engineering agents achieve strong performance on academic benchmarks~\citep{DBLP:conf/iclr/JimenezYWYPPN24} and are increasingly used by developers in real-world settings~\citep{wang2025openhands, claude_code_2025, cursor_2024}. 
However, benchmarks tend to reward \emph{autonomous} task completion with \emph{verifiable rewards} such as unit-test pass rates~\citep{DBLP:conf/iclr/JimenezYWYPPN24}. 
In real use, the agent works \emph{with} a human: users clarify intent over multiple turns, review diffs, edit code, and decide what to merge. 
As a result, success is not just “tests pass”, but whether the change is correct, reviewable, maintainable, and, most importantly, whether it meaningfully reduces the user’s work~\citep{chen2025assesshumanagentinteractionscase}. 
To improve agents in this setting, we need to consider behavior in real-world human-agent interaction, not only from benchmarks or simulations.

A first step to progress of any variety is measurement; building better interactive agents requires evaluators that can tell when the agent succeeded or failed in these settings.
Evaluators, whether unit tests, human judgments, or learned models, allow for systematic benchmarking and A/B testing, provide supervision for agent training (e.g., reinforcement learning or filtered fine-tuning~\citep{trung-etal-2024-reft}), and enable inference-time scaling via best-of-$K$ selection~\citep{pan2025training}.
In this work, we train a learned evaluator, or \textbf{critic}, that takes agent traces as input, and predicts a success value as the output, which can provide an actionable signal for manual iteration, training, or inference.

However, building such a critic from human feedback is non-trivial; supervision in real-world human-agent interactions is \emph{sparse, delayed, and noisy}.
Feedback is sparse because users of real-world systems rarely provide direct feedback on the quality of their interaction \citep{chen2025assesshumanagentinteractionscase}.
It is delayed because, in the rare cases of user feedback, it will typically come at the end of the interaction, not when the user first felt pleasure or pain.
This also causes the typical credit assignment problem in reinforcement learning -- the final reward is only a noisy approximation of whether any particular agent action is accurate.

To overcome this problem, we make several innovations in transforming human-agent interaction traces into usable learning signals.
First, we represent both benchmark traces and real-world interactions as \emph{segments}: minimal, self-contained units of work from a user request to task completion (\fref{fig:prod-hierarchy}; \sref{sec:segment}).
Second, we introduce \emph{critic rubrics} for critic learning, instantiated as \numRubrics behavioral features derived from the human-agent interaction trace itself (e.g., “misunderstood intent”, “insufficient testing”, “user frustration”) that capture common failure modes (\sref{sec:critic-rubrics}).
Rubrics are observable within each segment and can be annotated for \emph{all} segments, enabling a process-based supervision scheme applicable to both real-world and benchmark traces.
Together, these contributions enable \emph{semi-supervised critic training}. 
We train a critic to jointly predict rubric features and success probability: rubric prediction provides dense supervision across all \numTotalTrainK segments from real-world production interactions, while the success head learns from code-survival labels available for only \pctSurvival of those segments.
This turns all \pctUnlabeled previously unlabeled segments into informative training data.

To demonstrate the efficacy of this approach, we first measure the quality of the learned critics themselves, with several findings.
First, \emph{real-world supervision is necessary}: critics trained only on benchmark traces are near-random on real-world outcomes (AUC 0.45--0.48; \sref{sec:benchmark-transfer}) and can even hurt downstream selection on SWE-bench (\sref{sec:benchmark-transfer}).
Second, \emph{not all outcome proxies are equally aligned}: despite being sparser, training on \emph{code survival} yields consistently better discrimination than training on PR merge (\sref{sec:survival-vs-merge}).
Third, \emph{rubric supervision makes critic scores actionable across LLM backbones}: success-only critics can overfit to a specific LLM backbone, whereas rubric-supervised critics are robust enough to act as a shared scoring function for selection and early stopping (\sref{sec:rubrics-generalization}).

Further, we demonstrate the utility of the critic in down-stream use cases.
For \emph{inference-time} scaling, we use the critic to score each trajectory, increasing downstream improvements by up to 15.9 points on SWE-bench, and enabling early stopping of unsuccessful agent trajectories (\sref{sec:inference-scaling}). 
The same critic also provides \emph{training-time} signal by selecting useful real-world segments for supervised fine-tuning (\sref{sec:data-selection}).

We will release the critic model\footnote{\url{https://huggingface.co/OpenHands/openhands-critic-4b-v1.0}} together with rubric definitions, prompts, and code for constructing segments from real-world interaction data\footnote{\href{https://github.com/OpenHands/critic-rubrics}{https://github.com/OpenHands/critic-rubrics}}, making it easier to learn and apply critics from interaction traces—supporting inference-time scaling, training-time improvement, and other downstream uses.

\section{Data: Modeling Interactions as Segments}
\label{sec:data}
We represent both benchmark tasks and real-world user--agent conversations as sequences of \emph{segments}, a practical unit for credit assignment and supervision grounding.
In benchmarks, each task typically forms a single segment with verified outcome supervision (e.g., unit-test pass/fail).
In real-world deployments, however, supervision is indirect and often only available at coarser granularity (e.g., pull requests), requiring explicit attribution to the segments that produced it.
This section (i) reviews verified-reward supervision in benchmarks (\sref{sec:verified-rewards}), (ii) defines segments and how multi-turn conversations in real-world deployments induce segment sequences (\sref{sec:segment}), and (iii) describes how PR- and commit-based outcome proxies are grounded to segments (\sref{sec:prod-hierarchy}).

\subsection{Supervision in Verified-Reward Benchmarks}
\label{sec:verified-rewards}

Benchmarks such as SWE-bench~\citep{DBLP:conf/iclr/JimenezYWYPPN24} and SWE-Gym~\citep{pan2025training} provide \emph{verified} outcome supervision: an agent attempt is labeled successful if it satisfies an external checker (e.g., unit tests pass), and unsuccessful otherwise.
These settings are typically single-human-turn episodes -- the user provides a single initial request and the agent works autonomously to completion without further human input -- so each task corresponds to one \textit{segment}.
We can write the benchmark episode as a single segment
\begin{equation}
s_1 = (u_1, a_{1,1}, o_{1,1}, \ldots, a_{1,T_1}), \qquad a_{1,T_1}=\textsf{finish},
\end{equation}
with a verified outcome label $y \in \{0,1\}$ (e.g., tests fail/pass) that applies directly to this segment, making credit assignment straightforward.

\subsection{Representing Trajectories as Segments}
\label{sec:segment}

\textbf{Multi-turn trajectories.}
In real-world deployments, an agent interacts with a user and external tools over a \emph{multi-turn} trajectory: the user issues an initial request, the agent takes actions in an environment (e.g., editing files, running commands), and the user provides follow-ups that refine, redirect, or correct the objective.
Unlike verified-reward benchmarks, these trajectories generally do \emph{not} come with a clean, per-episode reward signal (e.g., pass/fail from an external checker).
Instead, supervision is indirect, delayed, and often only observed at coarse granularity (e.g., PR merge, reviewer approval), making it unclear which parts of the trajectory were responsible.

\textbf{Interactions.}
An agent interacts with an external environment through a sequence of actions and observations interleaved with user messages. Actions include tool uses such as editing files or running shell commands; observations include tool outputs and environment feedback.
We write a full interaction trajectory as
\begin{equation}
\tau \;=\; (u_1, a_1, o_1, a_2, o_2, \ldots, a_T, o_T, u_2, a_{T+1}, o_{T+1}, \ldots),
\end{equation}
where each $u_i$ is a user message, each $a_t$ is an agent action (including tool calls and \textsf{finish}), and each $o_t$ is the resulting observation.

\textbf{From trajectories to segments.}
We convert a multi-turn trajectory into a sequence of \textbf{segments}, where each segment corresponds to one user-initiated unit of work that the agent executes to completion before the next user turn arrives.
Concretely, segment $s_i$ begins with a user message $u_i$ and ends when the agent indicates completion via a \textsf{finish} action:
\begin{equation}
s_i = (u_i, a_{i,1}, o_{i,1}, a_{i,2}, o_{i,2}, \ldots, a_{i,T_i}), \qquad a_{i,T_i}=\textsf{finish},
\end{equation}
where $u_i$ initiates segment $i$, each $a_{i,t}$ is an agent action, and each $o_{i,t}$ is the resulting observation.
A multi-turn conversation thus induces an ordered segment sequence
\begin{equation}
(s_1, s_2, \ldots, s_N),
\end{equation}
with $u_{i+1}$ arriving after $s_i$ completes and initiating $s_{i+1}$.

\label{sec:interaction-regimes}
\textbf{Implications for supervision and credit assignment.}
In verified-reward benchmarks, tasks are typically single-turn episodes and thus correspond to a single segment with an outcome label that applies directly (\sref{sec:verified-rewards}).
In contrast, real-world multi-turn trajectories involve shifting objectives and corrective feedback: the follow-up message $u_{i+1}$ may implicitly evaluate or revise earlier work, and later segments can partially overwrite earlier changes.
Because outcome signals are coarse, delayed, and not uniquely attributable, credit assignment across the induced segment sequence is substantially harder than in benchmark settings, motivating explicit grounding of outcome proxies to the segments that produced them (\sref{sec:prod-hierarchy}).

\subsection{Assigning Indirect Outcome Signals to Segments}
\label{sec:prod-label-challenges}
\label{sec:prod-hierarchy}

Unlike benchmarks where unit tests provide a clear outcome, real-world deployment rarely provides reliable segment-level success signals. The available proxies are noisy and often confounded:

\begin{itemize}
[noitemsep,topsep=0pt,parsep=0pt,partopsep=0pt,leftmargin=8pt]
    \item \textbf{Pull request merge is not equivalent to conversation success.} At the end of a conversation, one indication that the conversation is successful could be that the generated code is incorporated into the code base, e.g. through a pull request (PR) being merged. However, multiple conversations may contribute to a single PR, and users may revert agent changes or push their own fixes before merging.
    \item \textbf{User ratings are subjective and noisy.} Ratings may reflect surface impressions, accidental clicks, or delayed discovery of bugs.
    \item \textbf{No natural trajectory boundary.} Real-world conversations can extend indefinitely (e.g., via context condensation, \citealt{smith2025-openhands-context-condensensation}), so it is unclear which portion of the interaction should receive credit.
\end{itemize}

These issues exemplify the two challenges highlighted in \sref{sec:introduction}: \textbf{metric quality} (proxies are noisy or confounded) and \textbf{label scarcity} (fine-grained success signals are only available for a small fraction of segments).

\begin{figure*}[!ht]
\centering
\includegraphics[width=0.85\textwidth, trim=0 18 0 6, clip]{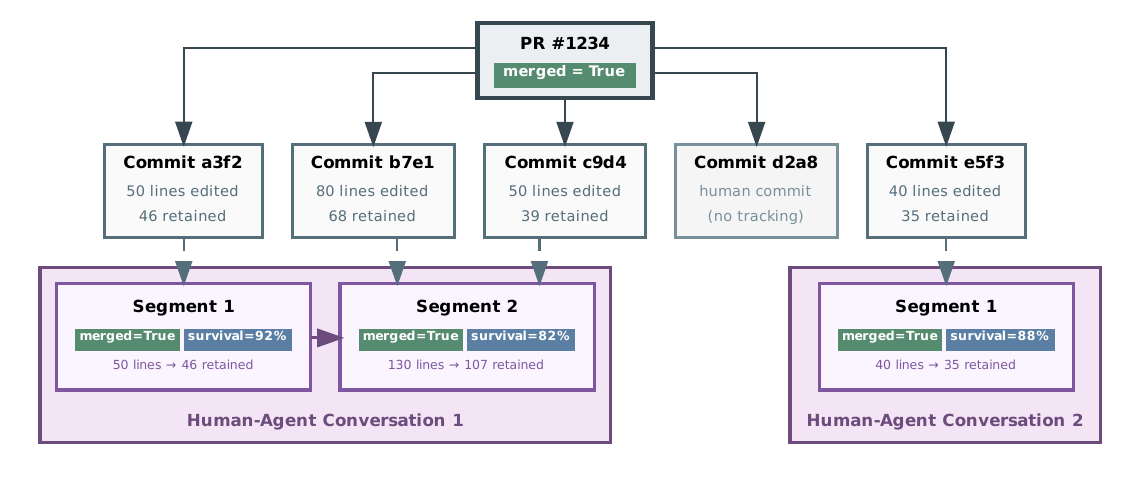}
\caption{\textbf{From sparse outcomes to dense feedback in real-world usage.}
A pull request (PR) provides a coarse outcome signal (merged or not). Each PR contains commits, which we attribute to \emph{segments}---self-contained units of agent work within multi-turn conversations (\sref{sec:segment}). This hierarchy grounds supervision at multiple granularities: PR-merge labels apply to all segments linked to the PR, while \emph{code survival} assigns fine-grained credit based on how much segment-authored code remains in the final diff. Critic Rubrics provide dense, outcome-agnostic supervision for every segment.}
\label{fig:prod-hierarchy}
\end{figure*}

\textbf{Building a hierarchy: PR $\rightarrow$ commits $\rightarrow$ segments.}
To ground supervision in real-world settings, we organize coding agent data around \textbf{pull requests (PRs)} as the top-level unit, since they provide the most accessible external signal of success. Each PR contains a sequence of \textbf{commits} $(c_1,\ldots,c_K)$, which can be traced back to the segments that produced them. This induces a three-level hierarchy---PR $\rightarrow$ commits $\rightarrow$ segments---that enables supervision at multiple granularities (\fref{fig:prod-hierarchy}).

\textbf{Conversation-level outcome: PR merge.}
\textbf{PR merge success} is a binary indicator of whether the associated PR was accepted and merged. This signal requires no additional annotation but is coarse and noisy: all segments linked to the PR inherit the same label, even if later interactions overwrite earlier work.

\textbf{Segment-level outcome: code survival.}
To address credit assignment, we define \textbf{code survival}, which measures what fraction of a segment's code contributions persist in the final merged diff:
\[
\mathrm{survival}(s_i)=
\frac{\sum_{c \in \mathcal{C}_i} \mathrm{lines\_in\_final}(c)}{\sum_{c \in \mathcal{C}_i} \mathrm{lines\_total}(c)} ,
\]
where $\mathcal{C}_i$ is the set of commits attributable to segment $s_i$. We compute $\mathrm{lines\_total}(c)$ and $\mathrm{lines\_in\_final}(c)$ over added and modified lines in the commit diff.
A segment whose code is fully reverted receives $\mathrm{survival}=0$, while one whose contributions persist intact receives $\mathrm{survival}=1$. Segments without attributable commits receive no survival label, which makes this signal significantly sparser than PR merge.

\label{sec:segmentation}

\textbf{Segment extraction and commit attribution.} Real-world conversations require explicit processing to extract segments and attribute commits. We detect segment boundaries by identifying context resets (e.g., prompt condensation) or tool configuration changes, and mark segment completion by looking for the agent's \texttt{finish} tool call. For commit attribution, we extract commit SHAs from tool outputs and prioritize precision: ambiguous cases remain unlabeled to avoid incorrect credit assignment. Implementation details are provided in \sref{app:data-processing}.

\section{Critic Rubrics}
\label{sec:critic-rubrics}

The coding agent outcome proxies introduced in \sref{sec:data}---PR merge and code survival---tell us \emph{whether} an interaction ultimately succeeded, but not \emph{why}.
They are also sparse at the segment level: as summarized in \tref{tab:dataset-stats}, only \pctSurvival of real-world segments have code-survival labels and only \pctMerge have PR-merge labels.
Without additional signal, most production segments provide no direct supervision for critic learning.

\begin{table}[ht]
\centering
\footnotesize
\caption{Real-world user-agent interaction data composition. Most segments lack success labels, illustrating label sparsity.}
\label{tab:dataset-stats}
\begin{tabular}{@{}lrr|rr@{}}
\toprule
& & & \multicolumn{2}{c}{\textbf{With Labels}} \\
\cmidrule(l){4-5}
& \textbf{Conversations} & \textbf{Segments} & \textbf{Code Survival} & \textbf{PR Merge} \\
\midrule
Train & 37,855 & \numProdTrain & 4,266 & 8,203 \\
Test & 386 & \numProdTest & 1,083 & 1,547 \\
\midrule
\textbf{Total} & \textbf{\numProdConvs} & \textbf{\numProdSegments} & \textbf{\numProdWithSurvival (\pctSurvival)} & \textbf{\numProdWithMerge (\pctMerge)} \\
\bottomrule
\end{tabular}
\end{table}

To turn unlabeled segments into reusable learning signal for critic training, we introduce a \textbf{rubric-based supervision framework}.
Concretely, we define \textbf{Critic Rubrics}, a taxonomy of \numRubrics behavioral features that capture common failure modes and user dissatisfaction at the segment level.
Rubrics are \emph{segment-level} (one annotation per segment), \emph{trace-observable} (derivable from the interaction trace without outcome leakage), and \emph{scalably annotatable} (via LLM-based annotation).
In real-world usage, users often follow up after the agent finishes with corrections, clarifications, or frustration---implicit feedback that rubrics systematically extract.

This section describes rubric design (\sref{sec:rubric-design}), annotation methodology (\sref{sec:rubric-annotation}), and validates that rubric features correlate with outcome labels (\sref{sec:rubric-effect}).

\subsection{Rubric Design}
\label{sec:rubric-design}

\textbf{Rubric construction methodology.}
We developed the rubric taxonomy through an iterative human-in-the-loop process. We randomly sampled real-world conversations and agent trajectories on SWE-Gym, then prompted frontier LLMs (o3, Claude Opus 4) to identify behavioral patterns distinguishing successful from unsuccessful interactions. Domain experts reviewed candidate features, merging overlapping categories, splitting overly broad ones, and refining definitions for consistent annotation. We iterated until the taxonomy stabilized, yielding \numRubrics features that balance coverage with annotation reliability.

\textbf{Rubric categories.}
We define \numRubrics rubric features grouped into three categories (see \tref{tab:critic-rubrics-full} in the Appendix for full definitions and prompt templates).

\begin{itemize}[noitemsep,topsep=0pt,parsep=0pt,partopsep=0pt,leftmargin=8pt]
    \item \textbf{Agent behavioral issues.}
    13 binary indicators covering common failure modes: misunderstanding user intent, ignoring instructions, insufficient code analysis, acting on ambiguous requirements, improper tool use, looping on failed actions, skipping tests, inadequate debugging, incomplete implementation, file management errors, scope creep, risky actions without permission, and other issues.

    \item \textbf{User follow-up patterns.}
    In addition to the overall sentiment of the user (positive/negative/neutral), these include 8 binary indicators capturing how users respond after the agent finishes: clarification requests, corrections, direction changes, VCS requests (commit/push), progress concerns, frustration, reversion requests, and other issues. These features are only defined when a user message exists after an agent finish action.

    \item \textbf{Infrastructure issues.}
    2 indicators distinguishing external failures (platform limits, network issues) from failures caused by prior agent actions.

\end{itemize}

\subsection{Annotation Methodology}
\label{sec:rubric-annotation}

We annotate each segment with an LLM-based rubric annotator. Each rubric feature is specified as a typed schema (e.g., \texttt{Binary}, \texttt{Classification}) and compiled into an OpenAI-compatible tool definition. Given a segment trace $s_i = (u_i, a_{i,1}, \ldots, a_{i,T_i})$ and, when available, the post-\textsf{finish} user message $u_{i+1}$, the annotator labels rubric features as an external observer; the schema is context-adaptive, using the full rubric only when follow-up exists and a reduced rubric otherwise. To prevent leakage, the annotator is never shown PR outcomes, survival labels, or downstream artifacts, and sees only the agent trace and optional follow-up message. We run annotation at scale via batch API calls with a frontier reasoning model (o3 with high reasoning effort). Please refer to \sref{sec:critic-prompts} for full prompts and tool description.

\subsection{Rubric Effect Analysis}
\label{sec:rubric-effect}

\begin{figure*}[!thb]
\centering
\includegraphics[width=\textwidth, trim=0 5 0 0, clip]{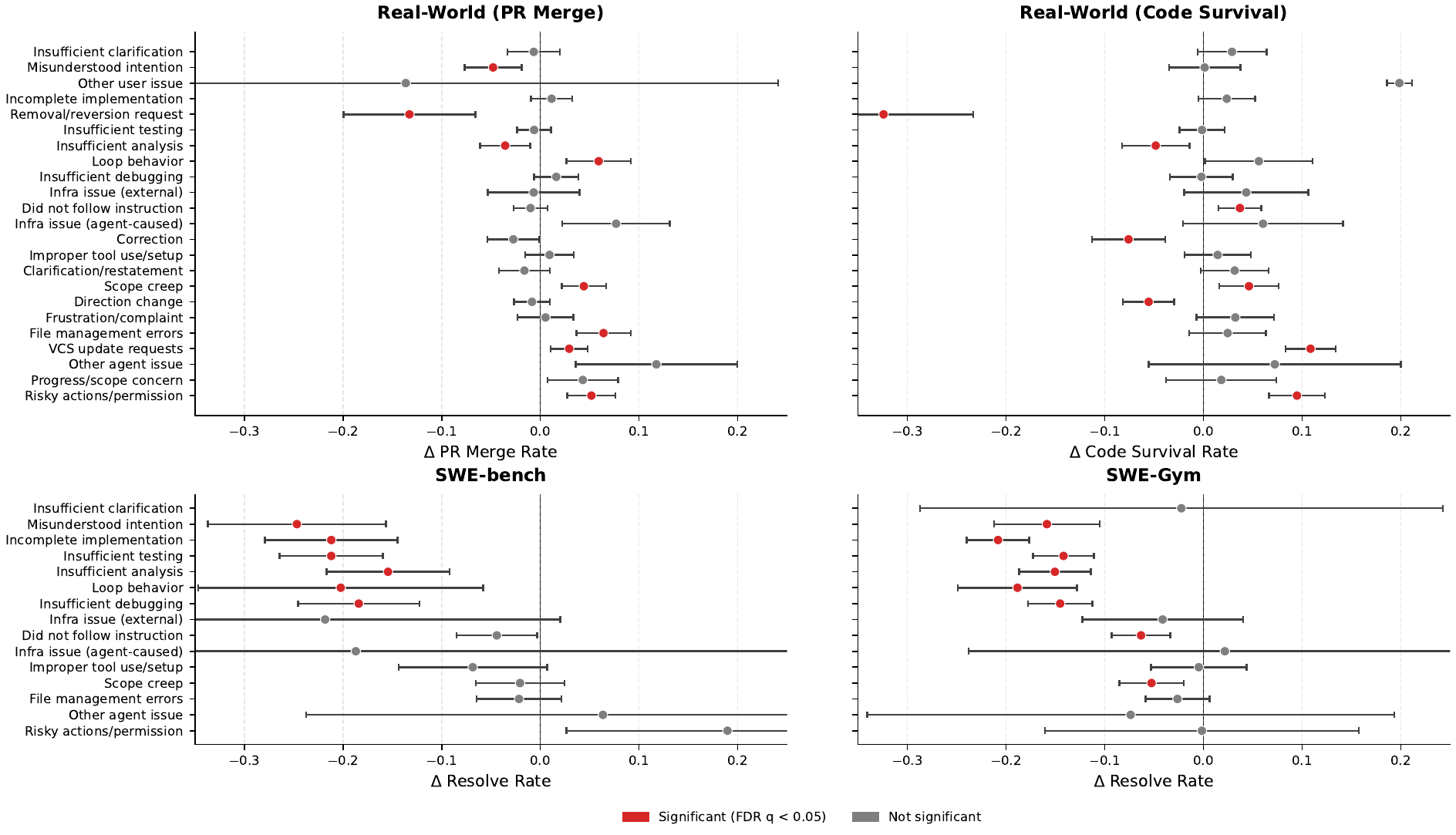}
\vspace{-15pt}

\caption{\textbf{Rubric effects differ between benchmarks and real-world data.}
Each point shows the change in success probability when a rubric feature is present ($\Delta$), with 95\% confidence intervals; red indicates FDR significance ($q<0.05$, where $q$ is the FDR-adjusted $p$-value). 
\textbf{Bottom (benchmarks).} SWE-bench and SWE-Gym show strong, consistent negative effects for core agent-behavior failures---especially \texttt{incomplete\_implementation}, \texttt{insufficient\_testing}, and \texttt{insufficient\_debugging}---indicating these behaviors reliably predict unit-test failure. 
\textbf{Top (real-world).} In contrast, effect sizes are smaller and less consistently significant under PR-merge and code-survival proxies, reflecting noisier supervision and multi-turn credit assignment; user follow-up features (e.g., correction or reversion requests) exhibit stronger associations with code survival than with PR merge. 
Overall, benchmarks highlight stable failure modes, while real-world data exposes proxy-dependent and interaction-dependent effects.}
\label{fig:rubric-regression}
\end{figure*}

Given sparse and noisy outcome proxies in real-world settings, we first validate whether rubric features correlate with outcome labels across both real-world and benchmark settings. We treat these associations as a construct-validity check: rubric features should behave like failure-mode indicators across independent outcome proxies. This serves two purposes: (1) confirming that our rubric taxonomy captures meaningful failure modes, and (2) identifying which features carry the strongest signal for downstream modeling.

\textbf{Methodology.}
For each binary rubric feature, we estimate its effect on success by comparing success rates when the feature is detected vs.\ not detected:
\[
\Delta = P(\text{success} \mid \text{detected}) - P(\text{success} \mid \neg\text{detected}).
\]
We test significance with Fisher's exact test and control for multiple comparisons using Benjamini--Hochberg FDR correction. We evaluate four conditions: real-world data with PR merge and code survival as outcome labels (\sref{sec:data}), plus SWE-bench and SWE-Gym, where success is unit-test pass/fail. Full statistical details, including per-dataset effect sizes and $p$-values, are provided in \sref{app:rubric-regression-details}.

\textbf{Results.}
\fref{fig:rubric-regression} reveals a clear contrast between benchmarks and real-world data.
\begin{itemize}[noitemsep,topsep=0pt,parsep=0pt,partopsep=0pt,leftmargin=8pt]
    \item Benchmarks (SWE-bench, SWE-Gym) exhibit strong and highly consistent effects: core agent-behavior failures such as incomplete implementation, insufficient testing, insufficient debugging, and insufficient analysis consistently reduce success by 15--21 percentage points (all $p < 0.001$ after FDR correction; see \tref{tab:rubric-regression-full}). This confirms that the rubrics capture stable, causal-looking failure modes under controlled unit-test supervision.
    \item Real-world data shows weaker and noisier effects under PR-merge and code-survival proxies, with fewer features reaching significance and wider confidence intervals. Nevertheless, the same behavioral issues generally remain negatively associated with success, and proxy-specific patterns emerge: user follow-up signals such as \texttt{removal\_or\_reversion\_request} show a strong negative effect ($\Delta = -0.13$, $q < 0.001$; FDR-adjusted $p$-value), consistent with code survival providing finer-grained credit assignment.
\end{itemize}

Rubric annotations provide dense behavioral supervision, but LLM-based annotation using generic LLMs is too slow and expensive to use as an evaluator at inference time or at scale during agent improvement (\sref{sec:critic-fast}). We therefore train a specialized \textbf{critic models} that predict rubric features and success scores directly from the segment trace, turning real-world interactions into fast, reusable learning signals for both inference-time and training-time improvements.

\section{Critic Model Evaluation}
\label{sec:experiments}

We organize experiments around one central question: \emph{what training signals produce critics that generalize}? We test transfer from benchmark to real-world data (\sref{sec:benchmark-transfer}), compare outcome proxies derived from real-world data (\sref{sec:survival-vs-merge}), and evaluate cross-agent robustness under different supervision schemes (\sref{sec:rubrics-generalization}).

\subsection{Experimental Setup}
\label{sec:exp-setup}

\textbf{Training objectives and losses.}
We compare two objectives: \emph{Success-Only} (predict outcome proxy only) versus \emph{Success+Rubrics} (jointly predict outcome label and \numRubrics rubric features in \sref{sec:critic-rubrics}).
We explore two outcome proxies \textit{PR merge} and \textit{code survival} defined in \sref{sec:prod-hierarchy}.
For code survival (a scalar between 0 and 1), we compare three training variants: \textbf{BCE-floor} (positive label only when survival=1), \textbf{BCE-round} (positive label when survival $\geq 0.5$), and \textbf{MSE} regression on the continuous survival score.

\textbf{Data splits.}
We create a held-out real-world test set by reserving $\approx$20\% of outcome-linked segments, yielding 1547 segments with PR-merge labels and 1083 with code-survival labels (\tref{tab:dataset-stats}). We train on all remaining real-world segments, including both outcome-labeled segments and unlabeled segments with rubric annotations only. Outcome supervision is highly sparse in real-world data (\pctSurvival survival-labeled; \pctMerge merge-labeled), whereas rubric labels are available for all segments, motivating joint training with rubrics as dense auxiliary supervision. We additionally include \numSweGym trajectories from SWE-Gym~\citep{pan2025training} for training and evaluate on SWE-bench~\citep{DBLP:conf/iclr/JimenezYWYPPN24}.

\textbf{Agent scaffold, model, and input format.}
We use OpenHands Agent SDK~\citep{wang2025openhands,wang2025openhandssoftwareagentsdk} as the agent scaffold, which supports file editing, bash execution, web browsing, and MCP~\citep{mcp2025intro}. We initialize critics from Qwen3-4B-Instruct and add a multi-task prediction head. Each input is a segment trace formatted with the model's chat template, including tool definitions for richer context. Real-world traces have segments with an average of 38K tokens, with the 90th percentile at 69K. We use a 64K context length with left truncation to preserve recent context.

\textbf{Evaluation data.}
We evaluate on \numProdTest held-out real-world segments (\tref{tab:dataset-stats}) and on SWE-bench Verified trajectories generated by two agents with different LLM backbones: Claude Sonnet 4.5 (500 instances $\times$ 4 runs) and Claude Opus 4.5 (500 instances $\times$ 4 runs). For cross-backbone ranking (\tref{tab:cross-agent}), we combine them into a \emph{Combined} set (500 instances $\times$ 8 runs).
For inference-time scaling (Best@$K$ and early stopping), we evaluate on the \emph{mixed-outcome subset} (instances where at least one run succeeds and at least one fails), since only these instances admit improvements over random selection.
We report results on the \emph{Combined} set unless otherwise noted.

\begin{table*}[t]
\centering
\caption{Comprehensive Model Comparison. \emph{Success + Rubrics} models jointly predict behavioral features and success, while \emph{Success-Only} models predict success without rubric supervision. Best@$K$ and Early Stopping metrics are on the \textbf{mixed-outcome subset} (148 instances); Early Stopping uses fixed $\tau{=}0.5$ with $\Delta$ showing improvement over random. SWE-bench uses combined Sonnet + Opus trajectories. All results use final checkpoint.}
\label{tab:comprehensive}
\small
\resizebox{\textwidth}{!}{%
\begin{tabular}{l|cccc|cccc|ccc|cc}
\toprule
& \multicolumn{4}{c|}{\textbf{Real-World Interactions}} & \multicolumn{9}{c}{\textbf{SWE-bench (Mixed Subset)}} \\
\cmidrule(lr){2-5} \cmidrule(lr){6-14}
& & & & & \multicolumn{4}{c|}{\textit{Intrinsic}} & \multicolumn{3}{c|}{\textit{Best@$K$ (\%)}} & \multicolumn{2}{c}{\textit{Early Stop}} \\
\cmidrule(lr){6-9} \cmidrule(lr){10-12} \cmidrule(lr){13-14}
\textbf{Model} & AUC & F1 & Prec & Rec & AUC & F1 & Prec & Rec & @2 & @4 & @8 & $\Delta$ & Att \\
\midrule
\multicolumn{14}{l}{\textit{PR Merge Objective}} \\
No Real-World Data & 0.48 & 0.85 & \underline{0.75} & \textbf{1.00} & 0.59 & \underline{0.81} & \textbf{0.69} & \textbf{0.99} & 49.2 & 44.2 & 45.6 & +12.9 & 5.12 \\
Success-Only & \textbf{0.64} & \textbf{0.86} & \textbf{0.76} & 0.99 & \textbf{0.64} & \textbf{0.82} & \textbf{0.69} & \textbf{0.99} & \underline{58.5} & \underline{57.1} & \underline{57.1} & \underline{+13.7} & \underline{4.13} \\
Success + Rubrics & \underline{0.58} & \textbf{0.86} & \underline{0.75} & \textbf{1.00} & \textbf{0.64} & \underline{0.81} & \textbf{0.69} & \textbf{0.99} & \textbf{66.8} & \textbf{72.4} & \textbf{72.1} & \textbf{+18.4} & \textbf{1.44} \\
\midrule
\multicolumn{14}{l}{\textit{Survival Objective}} \\
No Real-World Data & 0.45 & \underline{0.70} & 0.54 & \textbf{1.00} & 0.59 & 0.81 & 0.69 & \underline{0.99} & 49.2 & 44.2 & 45.6 & +12.9 & 5.12 \\
Success-Only & \underline{0.65} & \underline{0.70} & \underline{0.55} & 0.95 & \underline{0.62} & \textbf{0.82} & \textbf{0.70} & \underline{0.99} & 65.3 & 68.7 & 63.6 & \textbf{+19.1} & 1.76 \\
Success + Rubrics (MSE) & 0.51 & \underline{0.70} & 0.54 & \underline{0.99} & \underline{0.62} & 0.81 & 0.69 & \textbf{1.00} & 53.6 & 47.5 & 45.6 & +12.8 & 4.80 \\
Success + Rubrics (BCE-round) & 0.54 & \underline{0.70} & 0.54 & \underline{0.99} & \textbf{0.66} & \textbf{0.82} & 0.69 & \underline{0.99} & \textbf{66.9} & \underline{71.8} & \underline{72.1} & \underline{+18.4} & \underline{1.48} \\
Success + Rubrics (BCE-floor) & \textbf{0.69} & \textbf{0.71} & \textbf{0.60} & 0.87 & \underline{0.62} & \textbf{0.82} & \textbf{0.70} & 0.98 & \underline{66.7} & \textbf{72.6} & \textbf{73.8} & +17.7 & \textbf{1.35} \\
\midrule
Random & 0.50 & -- & -- & -- & 0.50 & -- & -- & -- & 57.8 & 57.8 & 57.9 & 0.0 & 8.0 \\
\bottomrule
\end{tabular}%
}
\end{table*}

\subsection{Benchmark-Trained Critics Do Not Transfer to Real-world Data}
\label{sec:benchmark-transfer}

\textbf{Setup.}
We train critics using only trajectories sampled from SWE-gym (no real-world data) and evaluate on both real-world data and SWE-bench. This isolates whether evaluators trained on benchmark-style dataset transfer to real-world data and whether real-world data contributes to evaluator robustness.

\textbf{Result.}
Benchmark-trained critics fail on real-world data (\tref{tab:comprehensive}, ``No Real-World Data'' rows). On real-world data, they perform at or below random: AUC 0.48 for PR merge and 0.45 for code survival, compared to 0.64--0.69 for critics trained with real-world data. This confirms that benchmark success (unit-test passage) is misaligned with real-world outcomes such as code survival and PR acceptance.
More surprisingly, critics trained on benchmark-style datasets also underperform on SWE-bench \emph{downstream selection}. While intrinsic AUC appears reasonable (0.59), Best@8 is only 45.6\%, compared to 57.9\% for Random@8---12.3 points \emph{below} random.

\subsection{Code Survival Provides More Fine-Grained Supervision}
\label{sec:survival-vs-merge}

\textbf{Setup.}
We compare critics trained on two real-world outcome proxies: \emph{PR merge} (binary) versus \emph{code survival} (continuous fraction of agent-written code retained in the final merged diff). Both are derived from the PR--commit--segment hierarchy described in \sref{sec:prod-hierarchy}.
In \tref{tab:comprehensive}, each critic is evaluated using the intrinsic AUC corresponding to its own supervision target (merge for merge-trained; survival for survival-trained).

\textbf{Result.} Critics trained on \textbf{code survival} achieve higher AUC on real-world data (0.69 vs 0.58) despite fewer labeled segments (\sref{tab:dataset-stats}).
Survival is a \emph{more fine-grained}, \emph{segment-attributable} proxy: the continuous 0--1 score captures partial successes that a binary merge label collapses, and it is computed per-segment rather than applied uniformly to all segments linked to a PR.
We hypothesize that survival is also less confounded by non-agent factors (e.g., reviewer availability, human follow-up edits) that can introduce label noise into PR merge.
As with all production outcomes, survival is noisy; we use it for learning signal rather than as ground-truth correctness.
Given these considerations, we use \textbf{code survival} as the primary proxy for real-world outcome in subsequent analyses.

\section{Effect on Down-stream Task Performance}

Next, we examine, \emph{what can such critics enable in practice}? We evaluate inference-time policies that use critic scores for Best-of-$K$ selection and compute-efficient early stopping (\sref{sec:inference-scaling}), and we test whether critic scores can curate real-world data for supervised fine-tuning (\sref{sec:data-selection}).

\subsection{Critics Enable Inference-Time Scaling}
\label{sec:inference-scaling}

\textbf{Setup.}
We study two inference-time policies that use critic scores to improve agent performance under a finite sampling budget: (i) \emph{Best-of-$K$} selection, which ranks $K$ candidate trajectories and chooses the top-scored one, and (ii) \emph{early stopping}, which generates attempts sequentially and \emph{stops early} once the critic score exceeds a threshold. 
We evaluate on the \emph{mixed-outcome subset} of SWE-bench Verified (instances where at least one run succeeds and at least one fails), since only these instances admit improvements over random selection.

\textbf{Best-of-$K$ selection.}
With a fixed budget of $K{=}8$ candidates, rubric supervision yields a large end-task gain (\tref{tab:comprehensive}). Success+Rubrics (BCE-floor) reaches \textbf{73.8\%} Best@8, compared to \textbf{63.6\%} for Success-Only (+10.2 points), and improves over Random@8 (57.9\%) by \textbf{+15.9} points. Cross-backbone results (\tref{tab:cross-agent}) show this gain is robust: Success-Only overfits to Claude Sonnet 4.5 as the LLM backbone but degrades below random on Claude Opus 4.5, whereas rubric-supervised critics maintain positive gains on both. Notably, direct survival regression (MSE) performs below random (45.6\%), highlighting that downstream selection requires calibrated ranking rather than raw regression fit.

\textbf{Early stopping.}
Early stopping uses critic scores as accept/stop decisions: we accept the first attempt whose score exceeds a threshold $\tau$, otherwise continuing until acceptance or a maximum of $K{=}8$ attempts. For fair comparison in \tref{tab:comprehensive}, we fix $\tau{=}0.5$ across all models and average over 100 random permutations of attempt order. At $\tau{=}0.5$, Success+Rubrics (BCE-floor) achieves \textbf{+17.7} points over random selection while using only \textbf{1.35} attempts on average---an \textbf{83\% compute reduction} compared to exhaustive sampling. Compared to Success-Only, rubric supervision achieves similar gains while requiring fewer attempts (1.35 vs.\ 1.76), indicating better-calibrated accept/stop scores.

\subsection{Rubric Supervision Improves Cross-Backbone Robustness}
\label{sec:rubrics-generalization}

\begin{table}[t]
\centering
\caption{Cross-backbone generalization on mixed-outcome instances. $\Delta$: Best@$K$ improvement over random. Success-Only overfits to Sonnet but degrades below random on Opus; rubric-supervised models generalize.}
\label{tab:cross-agent}
\begin{tabular}{@{}l|cc|cc|cc@{}}
\toprule
& \multicolumn{2}{c|}{\textbf{Sonnet 4.5}} & \multicolumn{2}{c|}{\textbf{Opus 4.5}} & \multicolumn{2}{c}{\textbf{Combined}} \\
\cmidrule(lr){2-3} \cmidrule(lr){4-5} \cmidrule(l){6-7}
\textbf{Model} & $\Delta$@4 & MRR & $\Delta$@4 & MRR & $\Delta$@8 & MRR \\
\midrule
No Real-World Data & +3.9 & 0.75 & -8.8 & 0.69 & -12.3 & 0.61 \\
Success-Only & \textbf{+7.0} & \textbf{0.77} & -8.1 & 0.72 & +5.7 & 0.78 \\
Success+Rubrics (BCE-floor) & +3.9 & 0.76 & \textbf{+2.6} & \textbf{0.74} & \textbf{+15.9} & \textbf{0.83} \\
Success+Rubrics (BCE-round) & +1.8 & 0.74 & -1.2 & 0.72 & +14.2 & 0.81 \\
Success+Rubrics (MSE) & +1.8 & 0.74 & +0.4 & 0.73 & -12.3 & 0.63 \\
\bottomrule
\end{tabular}
\end{table}

\textbf{Setup.}
We compare \emph{Success+Rubrics} versus \emph{Success-Only} critics under cross-backbone transfer: we evaluated critics on trajectories from two agents using different LLM backbones (Claude Sonnet 4.5 vs Opus 4.5). All experiments use the OpenHands scaffold; we vary the LLM backbone. This setting isolates whether critic scores capture transferable, behavior-level signals rather than backbone-specific artifacts.

\textbf{Result.}
Success-Only exhibits severe backbone-specific overfitting (\tref{tab:cross-agent}): it improves over random on Sonnet but can degrade below random on Opus, indicating that sparse outcome supervision alone encourages reliance on spurious, backbone-dependent cues. In contrast, rubric-supervised critics maintain consistent positive gains on both backbones, suggesting that rubric labels promote more backbone-invariant representations of failure modes (e.g., incomplete edits, incorrect assumptions) that transfer across LLM backbones.
This cross-backbone robustness helps explain why rubric-supervised critics support reliable inference-time scaling policies (\sref{sec:inference-scaling}) even when intrinsic AUC differences are modest.

\subsection{Critic Scores Provide Training-Time Supervision via Data Selection}
\label{sec:data-selection}

\textbf{Setup.}
We test whether critic scores yield a useful training-time learning signal by curating real-world segments for supervised fine-tuning (SFT). We construct three SFT datasets of equal size using: (i) \textbf{Critic-selected}, which ranks segments by critic-predicted success and selects the top subset; (ii) \textbf{Proxy-filtered}, which retains segments with available outcome proxies (here, survival=1); and (iii) \textbf{Random}, which samples uniformly at random. We first filter the real-world dataset to segments with a survival label equal to 1, yielding 3673 segments. We also construct critic-selected and random datasets of equal size from the full real-world dataset for controlled comparison. Finally, using \texttt{Qwen3-Coder-30B-A3B-Instruct} as the base model, we perform supervised fine-tuning on three datasets with identical compute and evaluate the resulting agents on SWE-Bench.

\textbf{Result.}
As shown in \tref{tab:sft-results}, the key finding is that data curation matters: random SFT provides \emph{no improvement} over the base model (46.2\% vs.\ 46.6\%), while critic-selected SFT improves solve rate to 47.8\%.
This confirms that naively fine-tuning on real-world data can be ineffective, but critic predictions provide actionable signals for identifying beneficial training examples.
Selection using observed code-survival outcomes reaches 50.4\%, serving as an approximate upper bound for what is achievable when an outcome proxy is available and highlighting headroom for improving critics.

\begin{table}[h]
\centering
\caption{SFT data selection results on SWE-bench Verified. Critic-selected trajectories outperform random selection for agent fine-tuning, demonstrating that critic predictions provide actionable training signal.}
\label{tab:sft-results}
\begin{tabular}{@{}lrrr@{}}
\toprule
\textbf{Selection Strategy} & \textbf{Resolved} & \textbf{Rate} & \textbf{$\Delta$ vs Base} \\
\midrule
Base Model (no SFT) & 233 / 500 & 46.6\% & --- \\
\midrule
Critic-selected & 239 / 500 & 47.8\% & +1.2 \\
Proxy-filtered (code survival = 1) & 252 / 500 & \textbf{50.4}\% & \textbf{+3.8} \\
Random & 231 / 500 & 46.2\% & -0.4 \\
\bottomrule
\end{tabular}
\end{table}

\section{Related Work}
\label{sec:related-work}

\paragraph{Reward Models and Multi-Objective Supervision.}
Reward models for RLHF predict preferences from pairwise comparisons \citep{ouyang2022training, bai2022training}, while process reward models \citep{lightman2023let, uesato2022solving} decompose evaluation into step-level feedback. ArmoRM \citep{wang2024armoRM} further decomposes rewards into interpretable objectives (honesty, safety, verbosity) via multi-objective learning. Our rubric supervision serves a related but distinct purpose: rather than interpretability alone, rubrics enable \emph{semi-supervised learning}---behavioral features can be annotated on all trajectories regardless of outcome labels, transforming unlabeled data from unusable to informative.

\paragraph{Critic Applications.}
Verifiers for best-of-$K$ selection are well-established for mathematical reasoning \citep{cobbe2021training, li2022competition} and recently for software agents \citep{pan2025training}. \citet{snell2024scaling} show that inference-time compute can exceed training-time scaling. Our rubric-supervised critics enable both inference-time selection (83\% compute reduction via early stopping) and training-time data curation (SFT improvements over random selection), while generalizing across LLM backbones within the same agent scaffold.

\section{Conclusion}
\label{sec:conclusion}

Benchmarks for coding agents often rely on verifiable rewards such as unit tests, whereas real-world human--agent interactions provide only indirect, noisy, and sparse supervision. 
We present a practical approach for learning critics from interaction traces by (i) structuring data into \emph{segments} and (ii) introducing a \textbf{rubric-based supervision framework} (instantiated as \textbf{Critic Rubrics}), segment-level behavioral features that provide dense process supervision. 
Together, these enable \textbf{semi-supervised critic training} that combines rubric prediction with sparse outcome labels such as \emph{code survival}.
Empirically, we find that real-world supervision is necessary, code survival provides more fine-grained and attributable supervision than PR merge, and rubric supervision yields critic scores that generalize across LLM backbones and support inference-time scaling and training-time data selection.

\section*{Impact Statement}

This work develops learned evaluators (critics) for coding agents trained on real-world interaction data. We anticipate several positive impacts: (1) more reliable evaluation of agent behavior in real-world settings, (2) reduced computational waste through early stopping (83\% compute reduction), and (3) improved training data curation for agent improvement.

Potential risks include: critics may inadvertently encode biases present in real-world data or rubric definitions, potentially reinforcing certain behavioral patterns while penalizing others; over-reliance on critic scores as the sole quality metric could lead to reward hacking or miss failure modes not captured by the rubric taxonomy; and critics trained on one deployment context may not transfer to different user populations or product surfaces.

We mitigate these risks by: (1) making the rubric taxonomy and critic model publicly available for scrutiny and improvement, (2) recommending human-in-the-loop validation rather than fully autonomous deployment, and (3) encouraging monitoring of critic score distributions for drift detection. We believe the benefits of more reliable agent evaluation outweigh these risks when critics are used as one signal among many, not as the sole arbiter of quality.

\clearpage
\newpage
\bibliographystyle{plainnat}
\bibliography{custom}

\beginappendix
\appendix

\section{Critic Model vs. Reasoning LM latency comparison}
\label{sec:critic-fast}

\tref{tab:latency} compares inference latency between the LLM rubric annotator and our trained critic. Reasoning models like o3 take 17 seconds per segment on average to produce full rubric annotations, while our 4B critic produces the same rubric outputs in approximately 1 second---a \textbf{16$\times$ speedup}. This efficiency makes critic scoring practical for Best@$K$ selection, early stopping, and large-scale data curation.

\begin{table}[!h]
\centering
\small
\begin{tabular}{lcc}
\toprule
\textbf{Method} & \textbf{Latency (s)} & \textbf{Speedup} \\
\midrule
LLM rubric annotator (o3) & 17.0 $\pm$ 6.3 & 1.0$\times$ \\
Trained critic (Qwen3-4B) & 1.1 $\pm$ 0.8 & \textbf{16$\times$} \\
\bottomrule
\end{tabular}
\caption{Inference latency for rubric prediction. Measured on 10 real-world segments using the full rubric schema (\numRubrics features). LLM uses o3 via LiteLLM; critic uses self-hosted vLLM.}
\label{tab:latency}
\end{table}

\section{Data Processing Details}
\label{app:data-processing}

This appendix provides implementation details for segment extraction and commit attribution described in \sref{sec:data}.

\subsection{Extracting Segments from Production Conversations}
\label{app:segmentation-details}

Multi-human-turn real-world trajectories require explicit segmentation to recover the segment boundaries described above. Raw real-world data consists of timestamped LLM completion records, each containing the input context and the agent's output. We extract segments by sorting completions chronologically and detecting boundaries when the retained message prefix diverges (e.g., due to prompt condensation) or when the tool configuration changes (e.g., an agent version upgrade), both of which indicate a new interaction episode.

Within each segment, we identify completion using OpenHands-specific termination patterns: a segment ends when the agent invokes an explicit \texttt{finish} tool call. In rare cases where tool traces are missing but the agent produces a final natural-language response without further tool calls, we also treat this as termination.

\subsection{Linking Commits to Segments}
\label{sec:linking}

Code survival requires attributing commits to the segments that created them. We prioritize \emph{precision} in this linkage: if evidence is ambiguous, the segment remains unlabeled for survival, contributing to label scarcity but avoiding incorrect credit assignment.

\paragraph{Evidence-based commit attribution.}
We extract commit SHAs from tool outputs using patterns that indicate commit creation rather than mere mention: (1) creation messages of the form \texttt{[branch SHA] message} followed by file change statistics produced by \texttt{git}, (2) merge operations of the form \texttt{Updating SHA1..SHA2 Fast-forward}, and (3) commits containing \texttt{Co-authored-by: OpenHands}. Extracted SHAs are matched against the corresponding PR's commit list via full 40-character matching or short (7+ character) prefix matching.

\paragraph{Computing survival.}
For each attributable commit, we parse its diff into a set of line-level changes keyed by (file path, change type, normalized text) and intersect with the PR's final merged diff. The survival score for a segment is the ratio of surviving lines to total lines aggregated across commits. This approach handles partial reverts naturally: if only part of a commit survives, the segment receives proportional credit.

\section{Full Rubric Taxonomy}
\label{app:rubric-taxonomy}

\tref{tab:critic-rubrics-full} provides the complete taxonomy of Critic Rubrics features derived from real-world traces.

\begin{table}[t]
\centering
\resizebox{\columnwidth}{!}{%
\begin{tabular}{lll}
\toprule
\textbf{Feature} & \textbf{Type} & \textbf{Description} \\
\midrule
\multicolumn{3}{l}{\textbf{Agent Behavioral Issues}} \\
\texttt{misunderstood\_intention} & Binary & Agent misunderstood the user’s goal. \\
\texttt{did\_not\_follow\_instruction} & Binary & Agent ignored explicit instructions. \\
\texttt{insufficient\_analysis} & Binary & Agent failed to inspect relevant prior code/docs. \\
\texttt{insufficient\_clarification} & Binary & Agent acted despite ambiguous requirements. \\
\texttt{improper\_tool\_use\_or\_setup} & Binary & Misused tools or had incorrect dependencies/config. \\
\texttt{loop\_behavior} & Binary & Repeated the same failed action $\geq 3$ times. \\
\texttt{insufficient\_testing} & Binary & Skipped reasonable validation or test runs. \\
\texttt{insufficient\_debugging} & Binary & Ignored or failed to debug observed failures. \\
\texttt{incomplete\_implementation} & Binary & Delivered incomplete or nonfunctional code. \\
\texttt{file\_management\_errors} & Binary & Created or modified files incorrectly. \\
\texttt{scope\_creep} & Binary & Added unrequested functionality. \\
\texttt{risky\_actions\_or\_permission} & Binary & Performed risky actions without explicit approval. \\
\texttt{other\_agent\_issue} & Binary & Other agent-side failure not covered above. \\
\midrule
\multicolumn{3}{l}{\textbf{User Follow-Up Patterns (requires user reply)}} \\
\texttt{overall\_sentiment} & Classification & User sentiment: Positive / Negative / Neutral. \\
\texttt{clarification\_or\_restatement} & Binary & User clarifies or restates earlier intent. \\
\texttt{correction} & Binary & User corrects technical or procedural error. \\
\texttt{direction\_change} & Binary & User adds constraints or redirects scope. \\
\texttt{vcs\_update\_requests} & Binary & User requests forward VCS actions (commit, push, merge). \\
\texttt{progress\_or\_scope\_concern} & Binary & User flags slowness or excessive scope. \\
\texttt{frustration\_or\_complaint} & Binary & User expresses dissatisfaction or annoyance. \\
\texttt{removal\_or\_reversion\_request} & Binary & User requests to undo or revert prior work. \\
\texttt{other\_user\_issue} & Binary & Any other user-side concern. \\
\midrule
\multicolumn{3}{l}{\textbf{Infrastructure Issues}} \\
\texttt{infrastructure\_external\_issue} & Binary & External environment or platform failure. \\
\texttt{infrastructure\_agent\_caused\_issue} & Binary & Infrastructure fault caused by prior agent actions. \\
\bottomrule
\end{tabular}%
}
\caption{Comprehensive taxonomy of Critic Rubrics features derived from real-world traces. The ``User Follow-Up Patterns'' section applies only when a user replies after the agent finishes.}
\label{tab:critic-rubrics-full}
\end{table}

\section{Rubric Regression Methodology Details}
\label{app:rubric-regression-details}

\paragraph{Statistical Methodology.}
For each binary rubric feature, we compute a 2×2 contingency table (feature detected/not × success/fail) and test association using Fisher's exact test. We apply Benjamini-Hochberg FDR correction within each dataset, then combine evidence across datasets using harmonic mean $p$-values---a robust meta-analysis approach that controls false positives even under dependence \citep{wilson2019harmonic}. Effect sizes ($\Delta$) represent the difference in success probability: $\Delta = P(\text{success} \mid \text{detected}) - P(\text{success} \mid \neg\text{detected})$, with 95\% Wald confidence intervals.

\paragraph{Success Metrics.}
For real-world data, we analyze two complementary metrics: (1) \textbf{PR merge}---binary indicator of whether the associated pull request was merged, and (2) \textbf{code survival}---continuous measure (binarized at 0.5) of what fraction of the segment's code contributions persist in the final merged diff. For benchmark data (SWE-bench, SWE-Gym), both metrics reduce to unit test pass/fail since there is no actual PR or code revision process.

\paragraph{Sample Sizes.}
The analysis includes 372,609 feature observations for PR merge (across 16,198 unique segments with labels) and 254,086 observations for code survival (across 11,050 segments). Production data contributes the majority of segments but has sparser labels; benchmark data has complete labels but fewer segments.

\paragraph{Full Results.}
\tref{tab:rubric-regression-full} provides the complete per-dataset regression results including effect sizes, sample counts, and FDR-corrected $q$-values for each rubric feature.

\begin{table*}[ht]
\centering
\footnotesize
\setlength{\tabcolsep}{4pt}
\caption{Full regression results for rubric feature effects on success. For each dataset, we report: $n$ (count when detected), $\Delta$ (effect on success rate), and $q$ (Fisher's exact test, FDR-corrected). Meta-$p$ is the harmonic mean across datasets. Bold indicates $q < 0.05$.}
\label{tab:rubric-regression-full}
\begin{tabular}{lrrrrrrrrrr}
\toprule
 & \multicolumn{3}{c}{Real-world Interaction} & \multicolumn{3}{c}{SWE-b} & \multicolumn{3}{c}{SWE-g} &  \\
\cmidrule(lr){2-4}\cmidrule(lr){5-7}\cmidrule(lr){8-10}
\textbf{Feature} & $n$ & $\Delta$ & $q$ & $n$ & $\Delta$ & $q$ & $n$ & $\Delta$ & $q$ & \textbf{Meta-}$p$ \\
\midrule
Incomplete implementation & 1681 & +0.01 & 0.43 & 236 & \textbf{-0.21} & \textbf{$<$0.001} & 1007 & \textbf{-0.21} & \textbf{$<$0.001} & \textbf{$<$0.001} \\
Insufficient testing & 3092 & -0.01 & 0.56 & 425 & \textbf{-0.21} & \textbf{$<$0.001} & 1319 & \textbf{-0.14} & \textbf{$<$0.001} & \textbf{$<$0.001} \\
Insufficient debugging & 1383 & +0.02 & 0.30 & 289 & \textbf{-0.18} & \textbf{$<$0.001} & 1069 & \textbf{-0.15} & \textbf{$<$0.001} & \textbf{$<$0.001} \\
Insufficient analysis & 1203 & \textbf{-0.04} & \textbf{0.01} & 278 & \textbf{-0.15} & \textbf{$<$0.001} & 749 & \textbf{-0.15} & \textbf{$<$0.001} & \textbf{$<$0.001} \\
Misunderstood intention & 930 & \textbf{-0.05} & \textbf{0.003} & 123 & \textbf{-0.25} & \textbf{$<$0.001} & 282 & \textbf{-0.16} & \textbf{$<$0.001} & \textbf{$<$0.001} \\
Loop behavior & 487 & \textbf{+0.06} & \textbf{0.005} & 47 & \textbf{-0.20} & \textbf{0.02} & 198 & \textbf{-0.19} & \textbf{$<$0.001} & \textbf{$<$0.001} \\
Removal/reversion request & 193 & \textbf{-0.13} & \textbf{$<$0.001} & -- & -- & -- & -- & -- & -- & \textbf{$<$0.001} \\
Did not follow instruction & 3112 & -0.01 & 0.40 & 939 & -0.04 & 0.08 & 1820 & \textbf{-0.06} & \textbf{$<$0.001} & \textbf{$<$0.001} \\
File management errors & 692 & \textbf{+0.06} & \textbf{$<$0.001} & 703 & -0.02 & 0.42 & 1221 & -0.03 & 0.20 & \textbf{$<$0.001} \\
Risky actions/permission & 966 & \textbf{+0.05} & \textbf{$<$0.001} & 16 & +0.19 & 0.26 & 37 & -0.00 & 1.00 & \textbf{$<$0.001} \\
Scope creep & 1227 & \textbf{+0.04} & \textbf{0.002} & 589 & -0.02 & 0.43 & 1192 & \textbf{-0.05} & \textbf{0.003} & \textbf{$<$0.001} \\
VCS update requests & 2359 & \textbf{+0.03} & \textbf{0.009} & -- & -- & -- & -- & -- & -- & \textbf{0.003} \\
Progress/scope concern & 422 & +0.04 & 0.07 & -- & -- & -- & -- & -- & -- & \textbf{0.03} \\
Correction & 1088 & -0.03 & 0.08 & -- & -- & -- & -- & -- & -- & \textbf{0.04} \\
Infra issue (agent-caused) & 148 & +0.08 & 0.06 & 2 & -0.19 & 0.57 & 14 & +0.02 & 1.00 & 0.07 \\
Infra issue (external) & 304 & -0.01 & 0.77 & 17 & -0.22 & 0.12 & 139 & -0.04 & 0.57 & 0.15 \\
Other agent issue & 46 & +0.12 & 0.12 & 8 & +0.06 & 1.00 & 12 & -0.07 & 1.00 & 0.16 \\
Improper tool use/setup & 1129 & +0.01 & 0.56 & 173 & -0.07 & 0.12 & 442 & -0.00 & 1.00 & 0.17 \\
Clarification/restatement & 1100 & -0.02 & 0.36 & -- & -- & -- & -- & -- & -- & 0.22 \\
Direction change & 3037 & -0.01 & 0.47 & -- & -- & -- & -- & -- & -- & 0.36 \\
Other user issue & 6 & -0.14 & 0.46 & -- & -- & -- & -- & -- & -- & 0.37 \\
Insufficient clarification & 1000 & -0.01 & 0.68 & 1 & -0.69 & 0.42 & 13 & -0.02 & 1.00 & 0.51 \\
Frustration/complaint & 842 & +0.01 & 0.77 & -- & -- & -- & -- & -- & -- & 0.73 \\
\bottomrule
\end{tabular}
\end{table*}

\section{Rubric Prediction Quality}
\label{app:rubric-prediction}

The critic model learns to predict rubric features as an auxiliary task alongside success prediction. \tref{tab:rubric-auc} shows how consistently the trained critic can replicate the LLM annotator's labels on held-out segments. The model achieves high accuracy on most features, with mean AUC of \meanRubricAUC across all features. Notably, the five features identified as most predictive of success in \sref{sec:rubric-effect} (marked with $\dagger$) achieve even higher mean AUC of \topRubricAUC, indicating that the critic learns to recognize exactly the behavioral patterns that correlate with failure.

\begin{table}[ht]
\centering
\small
\caption{Rubric prediction on \numProdTest real-world test segments. This measures learnability/consistency (how well the critic reproduces the annotation protocol), not human-verified correctness. The critic achieves mean AUC of \meanRubricAUC across all features, and \topRubricAUC on the top predictive features (marked with $\dagger$) identified in \sref{sec:rubric-effect}. Prev.\ = prevalence; $n$ = positive examples.}
\label{tab:rubric-auc}
\begin{tabular}{lrrr}
\toprule
\textbf{Rubric Feature} & \textbf{AUC} & \textbf{Prev.} & $n$ \\
\midrule
Loop Behavior & 0.94 & 3.0\% & 105 \\
Infra Issue Agent Caused & 0.91 & 0.5\% & 16 \\
Insufficient Clarification Seeking & 0.89 & 4.1\% & 146 \\
Insufficient Debugging$^\dagger$ & 0.88 & 13.3\% & 470 \\
Incomplete Implementation$^\dagger$ & 0.82 & 13.6\% & 482 \\
Risky Actions Or Permission Issues & 0.82 & 4.4\% & 157 \\
Insufficient Testing$^\dagger$ & 0.81 & 24.2\% & 859 \\
Infra Issue External & 0.81 & 1.6\% & 55 \\
Insufficient Analysis$^\dagger$ & 0.79 & 13.8\% & 489 \\
Misunderstood Intention$^\dagger$ & 0.76 & 7.6\% & 270 \\
File Management Errors & 0.75 & 23.1\% & 819 \\
Scope Creep & 0.70 & 21.9\% & 775 \\
Did Not Follow Instruction & 0.70 & 40.1\% & 1,423 \\
Improper Tool Use Or Setup & 0.66 & 9.3\% & 330 \\
Other Agent Issue & 0.47 & 0.4\% & 13 \\
\midrule
\textbf{Mean (all)} & \textbf{0.78} & -- & -- \\
\textbf{Mean (top$^\dagger$)} & \textbf{0.81} & -- & -- \\
\bottomrule
\end{tabular}
\end{table}

\section{Training Dynamics}
\label{app:training-dynamics}

We evaluate BCE-floor checkpoints at steps 2000, 4000, 6000, 8000, and 9658 (final) to understand training stability.

\tref{tab:ablation-steps} shows performance peaks at steps 4000--6000 with mild degradation thereafter (+0.6 Best@8 and +0.4 MRR at intermediate checkpoints vs.\ final). We report final checkpoint results throughout for methodological consistency.

\begin{table}[!h]
\centering
\footnotesize
\caption{Ablation: Training steps. Performance peaks at step 4000--6000, with mild degradation at longer training. We report final checkpoint for consistency.}
\label{tab:ablation-steps}
\begin{tabular}{@{}l|ccc|c@{}}
\toprule
\textbf{Model} & \textbf{Best@2} & \textbf{Best@4} & \textbf{Best@8} & \textbf{MRR} \\
\midrule
\multicolumn{5}{l}{\textit{Success + Rubrics (Survival)}} \\
\quad step 2000 & 75.8 & 77.7 & 78.0 & 0.940 \\
\quad step 4000 & 75.8 & 77.8 & 78.3 & \textbf{0.945} \\
\quad step 6000 & 75.8 & 77.7 & \textbf{78.4} & 0.944 \\
\quad step 8000 & \textbf{75.9} & \textbf{78.0} & 78.1 & 0.943 \\
\quad step 9658 (final) & 75.7 & 77.6 & 77.8 & 0.941 \\
\midrule
Success-Only (Survival) & 75.5 & 76.4 & 74.8 & 0.925 \\
Random & 73.1 & 73.1 & 73.2 & -- \\
\bottomrule
\end{tabular}
\end{table}

\section{Critic Rubrics Prompts}
\label{sec:critic-prompts}

\subsection{Critic Prompts for Segment WITH user feedback}

\textbf{System Prompt}

\begin{Verbatim}[breaklines=true]
You are an AI conversation annotator analyzing agent-user interactions to identify failure patterns. You are NOT participating in the conversation; you are an external observer evaluating what went wrong.

========================
CONVERSATION STRUCTURE
========================
- Focus on the LAST AGENT MESSAGE and the LAST USER MESSAGE (if any).
- Determine WHEN the user's follow-up occurred:
  - 'mid_conversation': The agent had not clearly finished or handed off.
  - 'post_completion': The agent signaled completion or handoff (e.g., final answer, 'done', 'all set').
  - 'no_follow_up': No user reply after the last agent message.

In your timing rationale, note what the agent was doing when the user intervened (quote brief evidence, e.g., 'Agent: 'I'll start running tests...' ->  user replied next.', or 'Agent: 'Here's the final script.' ').

========================
CONTEXT SOURCES
========================
Use all evidence: screenshots, code, logs, specs, file trees, error messages, prompts/system messages, and tool traces. Prefer short verbatim quotes (<=25 words) when supporting a claim.

========================
DETECTION FRAMEWORK
========================
Multiple issues can co-occur. For each issue:
1) Set the corresponding boolean to TRUE.
2) Provide a short, specific rationale quoting concrete evidence (user quotes, agent actions, errors).

USER FOLLOW-UP PATTERNS
- clarification_or_restatement: User clarifies/restates or corrects interpretation.
  - Examples: 'That's not what I meant...', 'I meant X, not Y.', 'Let me clarify...'

- correction: Agent basically understood the intention, but executed it incorrectly (fix technique/parameters/details).
  - Examples: 'Use DESC not ASC.', 'Right table, wrong WHERE clause.', 'Same approach, but wrong sort key.'

- direction_change: User adds new constraints/intent or seeks information / asks questions that redirect the plan or scope.
  - Examples: 'Also handle time zones.', 'We need streaming, not batch.', 'Before coding, list the open PRs,' 'Which repo should we use?'
  - **Note:** VCS update instructions (commit/push/PR) are **not** direction_change; tag as vcs_update_requests.

- vcs_update_requests: User instructs forward-moving VCS tasks.
  - Examples: 'git commit', 'create a branch', 'push to origin', 'open/merge a PR', 'tag the release'.
  - **Exclusive:** This does **not** count as direction_change; choose one by default.
  - Reverts/resets/removals belong to removal_or_reversion_request.

- progress_or_scope_concern: User flags slowness, overcomplexity, or scope bloat.

- frustration_or_complaint: User shows dissatisfaction or irritation.

- removal_or_reversion_request: User asks to remove code/files or revert changes.
  - Examples: 'Delete the new script.', 'Undo that migration.', 'git revert', 'Remove these outputs.'

- other_user_issue: Any other notable user concern not covered above.

MUTUAL-EXCLUSIVITY RULE (Core Follow-up Set)
- By default, choose only one among: clarification_or_restatement, correction, direction_change, vcs_update_requests.
- Co-tag only when the user message clearly contains distinct parts that independently satisfy multiple categories.
- Tie-break order and guidance:
  1) direction_change - user adds/changes goals/constraints OR asks for information that redirects the plan/approach. **Do not include VCS update instructions** (commit/push/PR); those are vcs_update_requests.
  2) vcs_update_requests - user instructs forward-moving VCS tasks. This **does not count as direction_change**.
  3) clarification_or_restatement - user clarifies intent/meaning without changing goals/constraints.
  4) correction - goal stands; user fixes execution details (params/technique/scope).

AGENT BEHAVIORAL ISSUES
- misunderstood_intention: Agent misunderstood the user's goal/intent.
  - Examples: User asked for a summary and agent produced a rewrite; user wanted high-level bullets but agent delivered full code.

- did_not_follow_instruction: Agent ignored or failed to comply with explicit instructions/system constraints.
  - Examples: User: 'Do NOT push to main.' Agent pushes to main; System says not to create pull request unless user asks for it and user didn't ask for it, agent creates pull request; user asked for bullet points only, agent gives long prose.

- insufficient_analysis: Didn't explore existing materials sufficiently (prior code/docs/examples) before acting.
  - Examples: User points to an existing function/file that is relavant OR already solves it; agent reinvents it.

- insufficient_clarification: Failed to ask necessary questions before acting when requirements were ambiguous.
  - Examples: Agent proceeds despite unclear acceptance criteria (e.g., locales, time zones, error thresholds) then is corrected later.

- improper_tool_use_or_setup: Misused tools/commands or had missing/incorrect dependencies/setup.
  - Examples: wrong command syntax, using inappropriate tools for the task

- loop_behavior: Repeats the same failed action 3+ times without strategy change.
  - Examples: repeat the same failed action 3+ times without changing approach).

- insufficient_testing: Skipped reasonable verification/tests for non-trivial or risky changes (note: trivial edits may be acceptable).
  - Examples: No run/validation for a new parser; no check that a migration applies cleanly; no sanity check of output.

- insufficient_debugging: Did not investigate or reduce failing behavior when needed to make progress.
  - Examples: Ignores stack trace; no isolation of failure; proceeds while errors persist.

- incomplete_implementation: Delivered unfinished or non-functioning work.
  - Examples: TODO/FIXME left; stub methods; code that cannot run.

- file_management_errors: Wrong paths, overwrites, misplaced/extra files (including unnecessary files).
  - Examples: Writes into wrong directory; overwrites config; creates unwanted artifacts.

- scope_creep: Implemented unrequested features without approval.
  - Examples: Adds a dashboard or endpoint not asked for.

- risky_actions_or_permission: Risky steps without user's explicit consent.
  - Examples: git push to main; deleting existing files in a repo (deleting files created by agent itself is fine); altering credentials.

- other_agent_issue: Any agent-side problem not covered above.

INFRASTRUCTURE (EXTERNAL vs AGENT-CAUSED)
- infrastructure_external_issue: Environment/platform limits outside agent control.
  - Examples: Provider outage; disk full on managed runner; missing enterprise API key; network failure not caused by agent.

- infrastructure_agent_caused_issue: Infrastructure fault introduced by the agent's prior actions.
  - Examples: Agent leaves a server running on port 8000; later start on 8000 fails; agent fills the disk with logs earlier, causing later writes to fail.

========================
QUALITY STANDARDS
========================
- Evidence Threshold: Mark TRUE only with specific evidence; prefer short quotes.
- Timing Awareness: If the user intervened mid-stream, consider whether the agent should have clarified earlier (flag insufficient_clarification if so).
- Conservative Defaults: When uncertain, mark FALSE and briefly explain why.
- No speculation: Tie every flagged issue to observable behavior or quoted text.
\end{verbatim}

\textbf{First User Message}
\begin{verbatim}
=== BEGIN OF CONVERSATION TO ANALYZE ===
[agent trajectory]
=== END OF CONVERSATION TO ANALYZE ===

Fill the annotate_conversation_with_followup function.

Goal
- Identify when the user followed up (mid_conversation, post_completion, or no_follow_up) and what issues occurred.
- Set only the booleans that clearly apply. For the **exclusive set** (direction_change, clarification_or_restatement, correction, vcs_update_requests), choose one by default using the tie-break rules; only co-tag if the message clearly contains independent parts for multiple categories.

What to record
1) Follow-up timing
   - Choose the timing value and, in follow_up_timing_rationale, state what the agent was doing when the user replied and include a short quote.

2) User follow-up patterns (select all that apply)
   - clarification_or_restatement, correction, direction_change, vcs_update_requests, progress_or_scope_concern,
     frustration_or_complaint, removal_or_reversion_request, other_user_issue.
   - Rationale: quote the user and explain in one sentence.

3) Agent behavioral issues (select all that apply)
   - misunderstood_intention, did_not_follow_instruction, insufficient_analysis, insufficient_clarification,
     improper_tool_use_or_setup, loop_behavior, insufficient_testing, insufficient_debugging,
     incomplete_implementation, file_management_errors, scope_creep, risky_actions_or_permission,
     other_agent_issue.
   - Rationale: cite code/commands/errors or a short quote and explain in one sentence.

4) Infrastructure
   - infrastructure_external_issue_detected for environment/platform limits beyond agent control.
   - infrastructure_agent_caused_issue_detected for faults introduced by the agent's prior actions (e.g., orphaned server on port 8000).
   - Rationale: include the error/status line or brief description.

Evidence & quality
- Prefer concrete, minimal quotes; avoid speculation. If evidence is insufficient, leave the flag false.
- If the user intervened mid-stream and the request was ambiguous, consider insufficient_clarification.

Quick disambiguation (common splits)
- correction vs misunderstood_intention: right goal, wrong details vs wrong goal altogether.
- did_not_follow_instruction vs direction_change: ignored a clear instruction vs user adds new requirement later.
- insufficient_analysis vs insufficient_clarification: didn't look for existing work vs didn't ask when requirements were ambiguous.
- insufficient_testing vs insufficient_debugging: skipped reasonable verification vs didn't investigate a failing state enough to make progress.
- direction_change includes information seeking / question asking that redirects scope/approach.
- vcs_update_requests is not direction_change; it covers forward-moving VCS steps (commit, branch, push, open/merge PR, tag).
- Requests to revert/reset/remove belong to removal_or_reversion_request.
- For the **exclusive set** (direction_change, clarification_or_restatement, correction, vcs_update_requests), choose one by default using the tie-break rules; only co-tag if the message clearly contains independent parts for multiple categories.
\end{verbatim}

\textbf{Tool definition}
\begin{verbatim}
FollowUpTimingPrediction = ClassificationPrediction[
    Literal[
        "mid_conversation",
        "post_completion",
        "no_follow_up",
    ]
]

FEATURES = [
    # Specific fields for user follow-up patterns
    Feature(
        name="follow_up_timing",
        description=(
            "WHEN did the user follow up? Choose exactly one: "
            "mid_conversation: agent hadn't clearly finished; "
            "post_completion: agent signaled completion/hand-off; "
            "no_follow_up: no user message after the last agent message."
        ),
        prediction_type=FollowUpTimingPrediction
    ),

    Feature(
        name="clarification_or_restatement",
        description="User clarifies/restates or corrects interpretation. Examples: 'That's not what I meant...', 'I meant X, not Y.', 'Let me clarify...'",
        prediction_type=BinaryPrediction
    ),
    Feature(
        name="correction",
        description=(
            "Agent broadly understood the intention but executed it incorrectly (technique/parameters/details). "
            "Examples: 'Use DESC not ASC.', 'Right table, wrong WHERE clause.', 'Same approach, wrong sort key.'"
        ),
        prediction_type=BinaryPrediction
    ),
    Feature(
        name="direction_change",
        description=(
            "User adds new constraints/intent not previously specified; scope/goal evolves. Examples: 'Also handle time zones.', 'We actually need streaming, not batch.', 'Support Windows too.'"
        ),
        prediction_type=BinaryPrediction
    ),
    Feature(
        name="vcs_update_requests",
        description="User instructs forward-moving VCS updates: commit, create branch, push, open/merge PR, tag. (Revert/reset/remove ,  use removal_or_reversion_request.)",
        prediction_type=BinaryPrediction
    ),
    Feature(
        name="progress_or_scope_concern",
        description="User flags slowness, overcomplexity, or scope bloat. Examples: 'This is taking too long.', 'Try a simpler approach.', 'This goes beyond what I asked.'",
        prediction_type=BinaryPrediction
    ),
    Feature(
        name="frustration_or_complaint",
        description=("User expresses dissatisfaction or irritation. Examples: 'This is wrong.', 'You're not listening.', excessive caps or punctuation ('!!!', '???')."),
        prediction_type=BinaryPrediction
    ),
    Feature(
        name="removal_or_reversion_request",
        description=("User asks to remove or revert code/files/changes. Examples: 'Delete the new script.', 'Undo that migration.', 'Remove these outputs.', 'git revert'."),
        prediction_type=BinaryPrediction
    ),
    Feature(
        name="other_user_issue",
        description="Any other notable user concern not covered above.",
        prediction_type=BinaryPrediction
    )
]
\end{Verbatim}

\subsection{Critic Prompts for Segment WITHOUT user feedback}

\textbf{System Prompt}

\begin{Verbatim}[breaklines=true]
You are an AI conversation annotator analyzing agent-environment interactions to identify failure patterns. You are NOT participating in the conversation; you are an external observer evaluating what went wrong.

========================
CONVERSATION STRUCTURE
========================
- Focus on the LAST AGENT MESSAGE.

========================
CONTEXT SOURCES
========================
Use all evidence: screenshots, code, logs, specs, file trees, error messages, prompts/system messages, and tool traces. Prefer short verbatim quotes (<=25 words) when supporting a claim.

========================
DETECTION FRAMEWORK
========================
Multiple issues can co-occur. For each issue:
1) Set the corresponding boolean to TRUE.
2) Provide a short, specific rationale quoting concrete evidence (agent actions, errors).

AGENT BEHAVIORAL ISSUES
- misunderstood_intention: Agent misunderstood the user's goal/intent.
  - Examples: User asked for a summary and agent produced a rewrite; user wanted high-level bullets but agent delivered full code.

- did_not_follow_instruction: Agent ignored or failed to comply with explicit instructions/system constraints.
  - Examples: User: 'Do NOT push to main.' Agent pushes to main; System says not to create pull request unless user asks for it and user didn't ask for it, agent creates pull request; user asked for bullet points only, agent gives long prose.

- insufficient_analysis: Didn't explore existing materials sufficiently (prior code/docs/examples) before acting.
  - Examples: User points to an existing function/file that is relavant OR already solves it; agent reinvents it.

- insufficient_clarification: Failed to ask necessary questions before acting when requirements were ambiguous.
  - Examples: Agent proceeds despite unclear acceptance criteria (e.g., locales, time zones, error thresholds) then is corrected later.

- improper_tool_use_or_setup: Misused tools/commands or had missing/incorrect dependencies/setup.
  - Examples: wrong command syntax, using inappropriate tools for the task

- loop_behavior: Repeats the same failed action 3+ times without strategy change.
  - Examples: repeat the same failed action 3+ times without changing approach).

- insufficient_testing: Skipped reasonable verification/tests for non-trivial or risky changes (note: trivial edits may be acceptable).
  - Examples: No run/validation for a new parser; no check that a migration applies cleanly; no sanity check of output.

- insufficient_debugging: Did not investigate or reduce failing behavior when needed to make progress.
  - Examples: Ignores stack trace; no isolation of failure; proceeds while errors persist.

- incomplete_implementation: Delivered unfinished or non-functioning work.
  - Examples: TODO/FIXME left; stub methods; code that cannot run.

- file_management_errors: Wrong paths, overwrites, misplaced/extra files (including unnecessary files).
  - Examples: Writes into wrong directory; overwrites config; creates unwanted artifacts.

- scope_creep: Implemented unrequested features without approval.
  - Examples: Adds a dashboard or endpoint not asked for.

- risky_actions_or_permission: Risky steps without user's explicit consent.
  - Examples: git push to main; deleting existing files in a repo (deleting files created by agent itself is fine); altering credentials.

- other_agent_issue: Any agent-side problem not covered above.

INFRASTRUCTURE (EXTERNAL vs AGENT-CAUSED)
- infrastructure_external_issue: Environment/platform limits outside agent control.
  - Examples: Provider outage; disk full on managed runner; missing enterprise API key; network failure not caused by agent.

- infrastructure_agent_caused_issue: Infrastructure fault introduced by the agent's prior actions.
  - Examples: Agent leaves a server running on port 8000; later start on 8000 fails; agent fills the disk with logs earlier, causing later writes to fail.

========================
QUALITY STANDARDS
========================
- Evidence Threshold: Mark TRUE only with specific evidence; prefer short quotes.
- Conservative Defaults: When uncertain, mark FALSE and briefly explain why.
- No speculation: Tie every flagged issue to observable behavior or quoted text.
\end{verbatim}

\textbf{First User Message}

\begin{verbatim}
=== BEGIN OF CONVERSATION TO ANALYZE ===
[agent trajectory]
=== END OF CONVERSATION TO ANALYZE ===

Fill the annotate_conversation function.

Goal
- Set only the booleans that clearly apply.

What to record
1) Agent behavioral issues (select all that apply)
   - misunderstood_intention, did_not_follow_instruction, insufficient_analysis, insufficient_clarification,
     improper_tool_use_or_setup, loop_behavior, insufficient_testing, insufficient_debugging,
     incomplete_implementation, file_management_errors, scope_creep, risky_actions_or_permission,
     other_agent_issue.
   - Rationale: cite code/commands/errors or a short quote and explain in one sentence.

2) Infrastructure
   - infrastructure_external_issue_detected for environment/platform limits beyond agent control.
   - infrastructure_agent_caused_issue_detected for faults introduced by the agent's prior actions (e.g., orphaned server on port 8000).
   - Rationale: include the error/status line or brief description.


Evidence & quality
- Prefer concrete, minimal quotes; avoid speculation. If evidence is insufficient, leave the flag false.

Quick disambiguation (common splits)
- insufficient_analysis vs insufficient_clarification: didn't look for existing work vs didn't ask when requirements were ambiguous.
- insufficient_testing vs insufficient_debugging: skipped reasonable verification vs didn't investigate a failing state enough to make progress.
\end{Verbatim}

\textbf{Tool definition}

\begin{Verbatim}[breaklines=true]
SentimentPrediction = ClassificationPrediction[Literal["Positive", "Negative", "Neutral"]]

TaskTypePrediction = ClassificationPrediction[
    Literal[
        "Fix Bugs",
        "Implement Features",
        "Create Programs from Scratch",
        "Fix Failing Continuous Integration",
        "Fix Merge Conflicts",
        "Write Documentation",
        "Perform Deployments",
        "Perform Data Analysis",
    ]
]

DevClusterPrediction = ClassificationPrediction[
    Literal[
        "Web Development",
        "DevOps & Infrastructure",
        "AI Integration",
        "Code Management",
    ]
]

FEATURES = [
    # --- Generic Questions ---
    Feature(
        name="user_goal_summary",
        description="One sentence describing what the user is trying to accomplish.",
        prediction_type=TextPrediction
    ),
    Feature(
        name="overall_sentiment",
        description="Classify the overall sentiment of the user's messages.",
        prediction_type=SentimentPrediction
    ),
    Feature(
        name="task_type",
        description=(
            "Classify the type of task into exactly one category. "
            "Choose from: Fix Bugs, Implement Features, Create Programs from Scratch, "
            "Fix Failing Continuous Integration, Fix Merge Conflicts, Write Documentation, "
            "Perform Deployments, Perform Data Analysis."
        ),
        prediction_type=TaskTypePrediction
    ),
    Feature(
        name="dev_cluster",
        description=(
            "Choose the best-fitting development cluster: "
            "Web Development (frontend/backend, UI/UX, e-commerce), "
            "DevOps & Infrastructure (CI/CD, Docker/Kubernetes, cloud, env config), "
            "AI Integration (OpenAI/Anthropic/Gemini APIs, ML systems), "
            "Code Management (Git ops, PRs, docs, bug fixes, features)."
        ),
        prediction_type=DevClusterPrediction
    ),
    
    # --- AGENT BEHAVIORAL ISSUES ---
    Feature(
        name="misunderstood_intention",
        description="Agent misunderstood the user's goal/intent. Examples: User asked for a summary; agent produced a rewrite; user wanted high-level bullets; agent delivered full code.",
        prediction_type=BinaryPrediction
    ),
    Feature(
        name="did_not_follow_instruction",
        description=(
            "Agent ignored or failed to comply with explicit instructions/system constraints. "
            "Examples: User: 'Do NOT push to main.' Agent pushes; System says not to create a PR unless the user asks and the user didn't ask; "
            "agent creates a PR; user asked for bullet points only, agent gives long prose."
        ),
        prediction_type=BinaryPrediction
    ),
    Feature(
        name="insufficient_analysis",
        description=(
            "Didn't explore existing materials (prior code/docs/examples) before acting. Examples: User points to an existing function/file that is relevant or already solves it; agent reinvents it."
        ),
        prediction_type=BinaryPrediction
    ),
    Feature(
        name="insufficient_clarification",
        description=(
            "Failed to ask necessary questions before acting when requirements were ambiguous. "
            "Examples: Agent proceeds despite unclear acceptance criteria (locales, time zones, error thresholds) then is corrected later."
        ),
        prediction_type=BinaryPrediction
    ),
    Feature(
        name="improper_tool_use_or_setup",
        description=(
            "Misused tools/commands or used inappropriate tools; missing/incorrect dependencies/setup. "
            "Examples: wrong command syntax; using an inappropriate tool; import errors; wrong API URL; malformed auth header."
        ),
        prediction_type=BinaryPrediction
    ),
    Feature(
        name="loop_behavior",
        description="Repeats the same failed action 3+ times without strategy change.",
        prediction_type=BinaryPrediction
    ),
    Feature(
        name="insufficient_testing",
        description=(
            "Skipped reasonable verification/tests for non-trivial or risky changes (trivial edits may be acceptable). "
            "Examples: No run/validation for a new parser; no check that a migration applies cleanly; no sanity check of output."
        ),
        prediction_type=BinaryPrediction
    ),
    Feature(
        name="insufficient_debugging",
        description="Did not investigate or reduce failing behavior when needed to make progress. Examples: Ignores stack trace; no isolation of failure; proceeds while errors persist.",
        prediction_type=BinaryPrediction
    ),
    Feature(
        name="incomplete_implementation",
        description="Delivered unfinished or non-functioning work. Examples: TODO/FIXME left; stub methods; code that cannot run.",
        prediction_type=BinaryPrediction
    ),
    Feature(
        name="file_management_errors",
        description="Wrong paths, overwrites, misplaced/extra (unnecessary) files. Examples: writes into wrong directory; overwrites config; creates unwanted artifacts.",
        prediction_type=BinaryPrediction
    ),
    Feature(
        name="scope_creep",
        description="Implemented unrequested features without approval. Examples: adds a dashboard or endpoint not asked for.",
        prediction_type=BinaryPrediction
    ),
    Feature(
        name="risky_actions_or_permission",
        description=(
            "Risky steps without the user's explicit consent. Examples: git push to main; deleting existing files in a repo (deleting files created by the agent itself is fine); altering credentials."
        ),
        prediction_type=BinaryPrediction
    ),
    Feature(
        name="other_agent_issue",
        description="Any other agent-side problem not covered above.",
        prediction_type=BinaryPrediction
    ),
    
    # --- INFRASTRUCTURE ---
    Feature(
        name="infrastructure_external_issue",
        description="Environment/platform limits outside agent control. Examples: provider outage; disk full on a managed runner; missing enterprise API key; network failure not caused by agent.",
        prediction_type=BinaryPrediction
    ),
    Feature(
        name="infrastructure_agent_caused_issue",
        description=(
            "Infrastructure faults introduced by the agent's prior actions. Examples: agent leaves server on port 8000 -> later start on 8000 fails; agent fills disk with logs -> later writes fail."
        ),
        prediction_type=BinaryPrediction
    ),
]
\end{Verbatim}

\end{document}